\documentclass[12pt]{article}
\usepackage{amsmath,ulem}
\usepackage{graphicx}
\usepackage{natbib}
\usepackage{url} 
\usepackage{amssymb, latexsym, bm, amsmath,amsthm, amsfonts,multirow, mathrsfs}
\usepackage{epsfig}
\usepackage{epstopdf}
\usepackage{algpseudocode}
\usepackage{algorithm}
\usepackage{subfigure}
\usepackage{capt-of}
\usepackage{color}

\def\Db{{\boldsymbol D}}

\def\Hb{{\boldsymbol H}}
\def\Ib{{\boldsymbol I}}

\def\Mb{{\boldsymbol M}}

\def\Xb{{\boldsymbol X}}

\def\betab{{\boldsymbol       \beta}}

\def\wb{{\boldsymbol w}}

\def\Gammab{{\boldsymbol      \Gamma}}

\def\Lambdab{{\boldsymbol     \Lambda}}

\def\Sigmab{{\boldsymbol      \Sigma}}

\graphicspath{{fig/}}

\newcommand{\blind}{0}

\newcommand{\argmin}{\mathop{\rm argmin}}

\def\1{{\bm 1}}
\def\0{{\bm 0}}
\def\th{{\rm th}}
\def\diag{{\rm diag}}

\def\Xb{\mathbf{X}}

\def\Mb{\mathbf{M}}
\def\Hb{\mathbf{H}}

\def\Db{\mathbf{D}}

\def\Ib{\mathbf{I}}

\def\wb{\mathbf{w}}
\def\Gammab{\mathbf{\Gamma}}

\def\det{{\rm det}}
\def\tr{{\rm tr}}

\newcommand{\real}{\mathbb{R}}
\renewcommand{\tilde}{\widetilde}
\renewcommand{\hat}{\widehat}

\def\Sr{{\mathcal{S}_r}}

\newtheorem{thm}{Theorem}
\newtheorem{prop}[thm]{Proposition}
\newtheorem{defn}[thm]{Definition}

\newtheorem{rmk}{Remark}
\newtheorem{cor}[thm]{Corollary}

\begin{document}

\def\spacingset#1{\renewcommand{\baselinestretch}%
{#1}\small\normalsize} \spacingset{1}


\if0\blind
{
  \title{\bf A Generalized Mean Approach for Distributed-PCA}
    \date{}
  \author{Zhi-Yu Jou,  Su-Yun Huang,\hspace{.2cm}\\
\small    Institute of Statistical Science, Academia Sinica, Taiwan\\
	Hung Hung\thanks{Corresponding author. {\it Email:} hhung@ntu.edu.tw},\\
\small Institute of Health Data Analytics and Statistics, National Taiwan University, Taiwan\\
    and \\
    Shinto Eguchi \\
\small    Institute of Statistical Mathematics, Japan}
  \maketitle

} \fi

\if1\blind
{
  \bigskip
  \bigskip
  \bigskip
  \begin{center}
    {\LARGE\bf Title}
\end{center}
  \medskip
} \fi

\bigskip
\begin{abstract}
Principal component analysis (PCA) is a widely used technique for dimension reduction. As datasets continue to grow in size,  distributed-PCA (DPCA) has become an active research area. A key challenge in DPCA lies in efficiently aggregating results across multiple machines or computing nodes due to computational overhead. Fan et al. (2019) introduced a pioneering DPCA method to estimate the leading rank-$r$ eigenspace, aggregating local rank-$r$ projection matrices by averaging. However, their method does not utilize eigenvalue information. 
In this article, we propose a novel DPCA method that incorporates eigenvalue information to aggregate local results via the matrix $\beta$-mean, which we call $\beta$-DPCA. The matrix $\beta$-mean offers a flexible and robust aggregation method through the adjustable choice of $\beta$ values. Notably, for $\beta=1$, it corresponds to the arithmetic mean; for $\beta=-1$, the harmonic mean; and as $\beta \to 0$, the geometric mean. Moreover, the matrix $\beta$-mean is shown to associate with the matrix $\beta$-divergence, a subclass of the Bregman matrix divergence, to support the robustness of $\beta$-DPCA. We also study the stability of eigenvector ordering under eigenvalue perturbation for $\beta$-DPCA. The performance of our proposal is evaluated through numerical studies.
\end{abstract}

\noindent%
\vfill

\newpage
\spacingset{1.75} 

\section{Introduction}

Let $X\in \mathbb{R}^p$ be a random vector of zero mean and covariance matrix $\Sigmab$. Consider the eigenvalue decomposition of $\Sigmab$ and its rank-$r$ truncated form: 
\begin{equation} 
\Sigmab = \Gammab \Lambdab \Gammab^{\top} \approx \Gammab_{r} \Lambdab_{r} \Gammab_{r}^{\top}, 
\end{equation} 
where $\Gammab \in \real^{p \times p}$ is an orthogonal matrix containing $p$ eigenvectors of $\Sigmab$, $\Gammab_{r}\in \real^{p \times r}$ represents the first $r$ columns of $\Gammab$, $\Lambdab \in \real^{p \times p}$ is a diagonal matrix with positive eigenvalues in descending order, and $\Lambdab_{r} \in \real^{r \times r}$ is the sub-matrix of $\Lambdab$ that consists of the $r$ largest eigenvalues. The goal of PCA is to estimate the leading rank-$r$ eigenspace $\Sr=\textrm{span}(\Gammab_{r})$ of $\Sigmab$ for a given $r$. Let 
$\Xb=[X_1,\ldots,X_n]$ be the $p \times n$ data matrix, where $X_i$'s are random copies of $X$. Then, PCA estimates $\Sr$ by the leading $r$ eigenvectors $\hat\Gammab_r$ of the sample covariance matrix $\hat\Sigmab=\frac{1}{n}\Xb\Xb^\top$. With the increasing size of modern datasets, computations are often distributed across multiple machines or computing nodes, leading to significant interest in distributed-PCA (DPCA). In DPCA, data are distributed across different machines or computing nodes, either by samples or by variables. This article focuses on sample partitioning, where each machine holds a complete set of variables for a subset of the samples. Let $\Xb = [\Xb^{(1)}, \Xb^{(2)}, \ldots, \Xb^{(m)}]$ represent the sample-partitioned data, where $\Xb^{(\ell)} \in \real^{p \times n_\ell}$ is the local data on machine $\ell$ with sample size $n_\ell$, and let $\widehat{\Sigmab}^{(\ell)} = \frac{1}{n_\ell} \Xb^{(\ell)} \Xb^{(\ell)^{\top}}$ be the sample covariance matrix of $\Xb^{(\ell)}$, $\ell=1,\ldots,m$. To minimize communication costs, rather than sending the entire local datasets to a central server, local estimators are computed on each machine and the results are sent to a central server for aggregation.

The idea of data partitioning for obtaining sub-results and then integrating to form the final result is not new. In the research filed of PCA, Yata and Aoshima (2010) first proposed the cross-data-matrix PCA (CDM-PCA), which corresponds to the case of $m=2$.
Bhaskara and Wijewardena (2019) proposed an algorithm that utilizes eigenvectors weighted by the square root of the eigenvalues. Li et al. (2021) introduced a communication-efficient distributed algorithm based on the power iteration method and the orthogonal Procrustes transformation to compute truncated SVD. A major contribution in this area was made by Fan et al. (2019), who proposed a DPCA algorithm that requires only one round of communication between local machines and the central server. Specifically, they aggregate sample rank-$r$ projection matrices from local machines by averaging, as summarized in Algorithm~\ref{algorithm_fan}. For a comprehensive review of DPCA methods, see Wu et al. (2018).

\begin{algorithm}[h]
 \caption{DPCA algorithm by Fan et al. (2019)}\label{algorithm_fan}
\begin{algorithmic}
 \vspace{0.25cm}
\State {With a prespecified rank $r$, go one round of the following procedure.}

\State \textbf{For local machine $\ell$:}
    \begin{itemize}
    \item[1.] Calculate the top $r$ eigenvectors $\hat{\Gammab}^{(\ell)}_{r}$ of the local sample covariance $\widehat{\Sigmab}^{(\ell)}$.

    \item[2.] Send $\hat{\Gammab}^{(\ell)}_{r}$ to the aggregator/central server.
    \end{itemize}

\State \textbf{For aggregator/central server:}
    \begin{itemize}
    \item[1.] Aggregate local projection matrices by averaging:
    \begin{equation} \label{aggregated_proj_matrix}
   \tilde{\Sigmab}= \frac{1}{m} \sum_{\ell=1}^{m} \hat{\Gammab}^{(\ell)}_{r} \hat{\Gammab}^{(\ell)^{\top}}_{r}.
     \end{equation}

    \item[2.] Calculate the leading $r$ eigenvectors $\tilde{\Gammab}_{r}$ of the aggregated matrix $\tilde{\Sigmab}$ in (\ref{aggregated_proj_matrix}).

    \end{itemize}
 \State \textbf{Output:} $\tilde{\Gammab}_{r}$.
\end{algorithmic}
\end{algorithm}

 While the algorithm by Fan et al. (2019) is communication-efficient, the eigenvalues of $\hat\Sigmab^{(\ell)}$'s are only utilized in the local machines to extract $\hat{\Gammab}^{(\ell)}_{r}$'s, but are not sent to the central server for aggregation. As a result, the aggregation in~(\ref{aggregated_proj_matrix}) depends solely on $\hat{\Gammab}^{(\ell)}_{r}$'s. From a statistical perspective, this can lead to less accurate estimation of $\Sr$. Another issue is that the aggregation result of DPCA can be sensitive to the failure of certain local results, potentially causing less accurate estimation of $\Sr$. By disregarding the eigenvalue information in~(\ref{aggregated_proj_matrix}), each $\widehat\Gammab_r^{(\ell)}$ is given equal weight, which makes Algorithm 1 vulnerable to the influence of biased local results. To address these limitations, we propose a novel DPCA algorithm that incorporates eigenvalue information during aggregation, utilizing the matrix $\beta$-mean to ensure robustness. The details of the matrix $\beta$-mean approach and the selection of a proper $\beta$ value will be presented in a later section. 

The remainder of this article is organized as follows. In Section~\ref{method}, we introduce our main method and the proposed algorithm for DPCA. Section~\ref{theoretical} presents theoretical properties to support our approach, linking it to a generalized matrix mean in Section~\ref{sec:beta_mean} and matrix divergence in Section~\ref{link_to_matrix_beta}. In Section~\ref{eig_perturb}, we examine the stability of eigenvector ordering under eigenvalue perturbation. Section~\ref{numerical} demonstrates the performance of our algorithm through simulation studies. The article concludes in Section~\ref{conclusion}, with technical details and proofs provided in the Appendix.

\section{Method}\label{method}

Our primary goal is to estimate the leading rank-$r$ eigenspace $\Sr$ in a distributed setting. For simplicity, assume that $n$ samples are evenly distributed across $m$ machines, with each local dataset represented as $\Xb^{(\ell)} \in \real^{p \times n_\ell}$, where $n_\ell = \frac{n}{m}$ for $\ell = 1, 2, \dots, m$. This equal allocation simplifies the presentation but does not restrict the general applicability of our method, which remains valid for unequal subsample sizes. While the approach of Fan et al. (2019) effectively aggregates sample rank-$r$ projection matrices without incorporating eigenvalue information, we propose to incorporate eigenvalues in the aggregation step.

\subsection{DPCA via $\bm\beta$-mean aggregation}

Each local machine is tasked with estimating the leading rank-$q$ eigenvalues and eigenvectors, where $q \ge r$ represents the sampling dimension, with $q - r$ accounting for the additional dimensions beyond the target rank $r$. 
Each machine transmits the local estimates $(\hat\Lambdab_{q}^{(\ell)}, \hat\Gammab_{q}^{(\ell)})$ to the central server. At the central server, we propose to aggregate the local results via the matrix $\beta$-mean (see Section~\ref{sec:beta_mean} for the detailed construction of matrix $\beta$-mean) as summarized below:
\begin{equation}\label{integrated_matrix}
\widehat{\Sigmab}_{\beta} =     \left\{\begin{array}{ll}
\left\{ \frac{1}{m} \sum_{\ell=1}^{m} \widehat{\Gammab}^{(\ell)}_{q}   \left(\widehat{\Lambdab}^{(\ell)}_{q}\right)^{\beta} 
   \widehat{\Gammab}^{(\ell)^{\top}}_{q}   \right\}^{1/\beta}, & {\rm for}~ \beta> 0, \\[1ex]
  \exp\left\{\frac{1}{m} \sum_{\ell=1}^m  \widehat{\Gammab}^{(\ell)}_{q} \ln \left( \widehat{\Lambdab}^{(\ell)}_{q}\right) \widehat{\Gammab}^{(\ell)^{\top}}_{q}\right\},  & {\rm for}~ \beta\to 0, \\[1ex]
     \left\{ \frac{1}{m} \sum_{\ell=1}^{m}  \left(\widehat{\Gammab}^{(\ell)}_{q} \widehat{\Lambdab}^{(\ell)}_{q} \widehat{\Gammab}^{(\ell)^{\top}}_{q} +  \delta     \Ib_{p} \right)^{\beta} \right\}^{1/\beta}, & {\rm for}~ \beta<0,
\end{array}\right.
\end{equation}
where $\beta$ is a tuning parameter controlling the performance of aggregation (the selection of $\beta$ will be discussed separately), and $\delta\Ib_p$ with $\delta >0$ is a regularization term for $\beta < 0$. We then propose to estimate $\Sr$ by the leading $r$ eigenvectors $\widehat{\Gammab}_{r}$ of the aggregated sample covariance $\widehat{\Sigmab}_{\beta}$, which we call $\beta$-DPCA. The full implementation of $\beta$-DPCA algorithm is summarized in Algorithm~\ref{algorithm_disPCA}.

\begin{algorithm}[h!]
 \caption{$\beta$-DPCA algorithm}\label{algorithm_disPCA}
\begin{algorithmic}
 \vspace{0.25cm}
\State {With a prespecified target rank $r$, an eigenvector sampling dimensionality $q\ge r$, and a tuning parameter $\beta$, go one round of the following procedure.}

\State \textbf{(L) For local machine $\ell$:}
    \begin{itemize}
    \item[1.]   Calculate the top $q$ eigenvectors $\hat{\Gammab}^{(\ell)}_{q}$ of the local sample covariance $\widehat{\Sigmab}^{(\ell)}$.

    \item[2.] Send $(\widehat{\Gammab}^{(\ell)}_{q}, \widehat{\Lambdab}^{(\ell)}_{q})$ to aggregator/central server.
    \end{itemize}

\State {\bf (C) For aggregator/central server:}
    \begin{itemize}
    \item[1.] Aggregate the local truncated PCA to get $\widehat{\Sigmab}_{\beta} $ given in~(\ref{integrated_matrix}).

    \item[2.] Calculate the leading $r$ eigenvectors $\widehat{\Gammab}_{r}$ of the matrix $\widehat{\Sigmab}_{\beta} $.
    \end{itemize}
 \State \textbf{Output:} $\widehat{\Gammab}_{r}$.
\end{algorithmic}
\end{algorithm}

The following three special cases of $\beta$-DPCA are of major interest, which can also be linked to existing methods:
\begin{itemize}
\item
For $\beta=1$, $\hat\Sigmab_\beta$ corresponds to taking average (arithmetic mean) of local truncated PCAs, i.e.,
$\frac{1}{m} \sum_{\ell=1}^{m}\widehat{\Gammab}^{(\ell)}_{q} \widehat{\Lambdab}^{(\ell)}_{q} \widehat{\Gammab}^{(\ell)^{\top}}_{q}$.

\item
For $\beta=-1$, ${\hat\Sigmab}_\beta$ corresponds to taking harmonic mean of local truncated PCAs, i.e., $\left\{\frac{1}{m} \sum_{\ell=1}^{m}\left(\widehat{\Gammab}^{(\ell)}_{q} \widehat{\Lambdab}^{(\ell)}_{q} \widehat{\Gammab}^{(\ell)^{\top}}_{q}+\delta\Ib_p \right)^{-1}\right\}^{-1}$.

\item
For $\beta\to 0$, ${\hat\Sigmab}_\beta$ corresponds to taking geometric mean of local truncated PCAs,  i.e.,
$\exp\left(\frac{1}{m} \sum_{\ell=1}^{m}\widehat{\Gammab}^{(\ell)}_{q}
\ln\big(\widehat{\Lambdab}^{(\ell)}_{q}\big) \widehat{\Gammab}^{(\ell)^{\top}}_{q}\right)$.
\end{itemize}

\noindent  The method by Bhaskara and Wijewardena (2019) incorporates eigenvalue information by weighting the eigenvectors with the square root of eigenvalues, which corresponds to $\beta$-DPCA with $\beta = 1$. The method by Fan et al. (2019) assigns equal weights to the leading~$r$ eigenvalues for aggregation, which corresponds to the case of $\beta=0$. The limiting case of $\beta \to 0$ corresponds to taking the natural logarithm of the eigenvalues, which reduces to the CDM-PCA of Yata and Aoshima (2010) with $m=2$. 

\begin{rmk}
Similar to Fan et al. (2019), our method requires only one round of communication between each local machine and the central server. Except for the case of $\beta=0$, our algorithm incurs higher communication costs due to the need to transmit a larger eigenvector matrix of size $p \times q$ (instead of $p \times r$) and an additional cost for sending eigenvalues ($q \times 1$). 
\end{rmk}

\subsection{Selection of $\betab$}

Since different $\beta$ values lead to different results and distinct behavior of the results for the proposed method in Algorithm~\ref{algorithm_disPCA}, we design a selection procedure based on $K$-fold cross-validation to determine the optimal value of $\beta$.
The procedure is performed under the assumption of homogeneity among $m$ local machines.
Let $\mathcal{B}$ be a candidate set of the tuning parameter $\beta$.
We partition $m$ machines into disjoint $K$ folds. For $j=1, 2, \ldots, K$, the $j^\th$ fold is treated as the validation set $\mathcal{M}_{\textrm{valid}}^{(j)}$, and the remaining $K-1$ folds belong to the training set $\mathcal{M}_{\textrm{train}}^{(j)}$.
Hence, $| \mathcal{M}_{\textrm{valid}}^{(j)}| = \lfloor m/K \rfloor$ and $| \mathcal{M}_{\textrm{train}}^{(j)} | = m- \lfloor m/K \rfloor$ for $m>K$. If $m \leq K$, we recommend using leave-one-out cross-validation by re-setting $K = m$.
The detailed cross-validation procedure for our proposal is summarized in Algorithm~\ref{algorithm_cv}.

\begin{algorithm}[h!]
 \caption{Selection of $\beta$ by $K$-fold cross validation}\label{algorithm_cv}
\begin{algorithmic}
 \vspace{0.25cm}
\State Given the candidate set $\mathcal{B}$, go through the following procedure $j =1, 2, \ldots, K$.

\State \textbf{For local machine $\ell$ in $\mathcal{M}_{\textrm{train}}^{(j)}$:}
\begin{itemize}
\item
Apply Algorithm~\ref{algorithm_disPCA} step-(L) to local clients in training fold:
Calculate the leading~$q$ eigenvalues $\widehat{\Lambdab}^{(\ell)}_{q}$ and eigenvectors $\widehat{\Gammab}^{(\ell)}_{q}$ and send them to the central server.
    \end{itemize}

\State \textbf{For local machine $\tilde\ell$  in $\mathcal{M}_{\textrm{valid}}^{(j)}$:}
\begin{itemize}
\item
Apply Algorithm~\ref{algorithm_disPCA} step-(L) to local clients in validation fold:
Calculate the leading $r$ eigenvectors $\widehat{\Gammab}^{(\tilde\ell)}_{r}$ and send them to the central server.
 \end{itemize}

\State \textbf{For aggregator/central server:}
    \begin{itemize}
    \item Apply Algorithm~\ref{algorithm_disPCA} step-(C) to aggregate information from the $j^\th$ training fold $\mathcal{M}_{\textrm{train}}^{(j)}$ to get
    the leading rank-$r$ eigenvectors, denoted by $\widehat{\Gammab}_{r,\beta}^{(j)}$.

    \item Calculate the following discrepancy using validation set $\mathcal{M}_{\textrm{valid}}^{(j)}$:
    \[
        d^{(j)}(\beta) = \frac1{| \mathcal{M}_{\textrm{valid}}^{(j)}|} \sum_{ \tilde\ell \in \mathcal{M}_{\textrm{valid}}^{(j)}}
        \left\| \widehat{\Gammab}_{r,\beta}^{(j)} \widehat{\Gammab}_{r,\beta}^{(j)\top}
        - \widehat{\Gammab}^{(\tilde\ell)}_{r} \widehat{\Gammab}^{(\tilde\ell)^{\top}}_{r}\right \|_{F}^{2},
        ~~ | \mathcal{M}|:\mbox{cardinality of } \cal{M}.
    \]
    \item Find $\widehat{\beta} = \underset{\beta \in \mathcal{B}} \argmin \frac{1}{K} \sum_{j=1}^{K} d^{(j)}(\beta) $.
\end{itemize}

 \State \textbf{Output:} $\widehat{\beta}$.
\end{algorithmic}
\end{algorithm}

\pagebreak

\section{Theoretical Properties}\label{theoretical}

This section presents the theoretical support for the superiority of $\beta$-DPCA, via the connection to the matrix $\beta$-mean and the minimum matrix $\beta$-divergence estimation criterion. 


\subsection{Connection to matrix $\beta$-mean} \label{sec:beta_mean}

The proposed $\beta$-DPCA employs the idea of matrix $\beta$-mean for aggregation. The matrix-$\beta$ mean is a special case of the generalized matrix mean introduced by Eguchi and Komori (2022) as described below. Let $\cal P$ be the set of all symmetric positive
semi-definite matrices of size $p \times p$, and let $\phi:  \real^+ \to \real$ be a strictly increasing function. Given a set of matrices $\left\{\Mb^{(\ell)}\right\}_{\ell=1}^{m}$ in $\mathcal{P}$, the $\phi$-generalized mean of $\left\{ \Mb^{(\ell)}\right\}_{\ell=1}^{m}$ is defined to be
\begin{equation}\label{phi_generalized_mean}
 \overline{\Mb}_{\phi} = \phi^{-1} \left( \frac{1}{m} \sum_{\ell=1}^{m} \phi \left(\Mb^{(\ell)} \right) \right),
\end{equation}
where ``$\phi(\Mb^{(\ell)})$'' is understood as
\[\phi(\Mb^{(\ell)}) = \sum_{j=1}^p \phi(\lambda_j^{(\ell)}) \gamma_j^{(\ell)}\gamma_j^{^{(\ell)}\top} \]
with $\lambda_j^{(\ell)}$'s and $\gamma_j^{(\ell)}$'s being the eigenvalues and eigenvectors of $\Mb^{(\ell)}$.

Below, we introduce a special class of $\phi$ functions, the $\beta$-power functions (Eguchi and Kato 2010; Eguchi et al. 2018), to construct the matrix $\beta$-mean.

\begin{defn}[Matrix $\beta$-mean]
Let $\left\{\Mb^{(\ell)}\right\}_{\ell=1}^{m}$ be a set of matrices in $\cal P$.
The matrix $\beta$-mean (or simply $\beta$-mean for short) of $\left\{\Mb^{(\ell)}\right\}_{\ell=1}^{m}$ is defined to be the $\phi$-generalized mean with $\phi(t) = (t^{\beta} -1)/\beta$, $\beta \in \real \setminus\{0\}$:
\begin{equation}\label{phi_beta_mean}
 \overline{\Mb}_{\beta} =  \left\{ \frac{1}{m} \sum_{\ell=1}^{m}  \left(\Mb^{(\ell)}  \right)^{\beta} \right\}^{1/\beta}.
\end{equation}
Notably, when $\beta < 0$ and $\Mb^{(\ell)}$ is singular, we need to add a regularization $\delta\Ib_p$ with $\delta >0$ to $\Mb^{(\ell)}$.
The scaling factor $1/\beta$ and the term $-1$  in $\phi$ are for the limiting case, $\beta\to 0$. For other $\beta$ values, one can simply use $\phi(t)=t^\beta$.
\end{defn}

\begin{rmk}[limiting case of $\beta\to 0$]  \label{limiting_case_g_mean}
 When $\beta\to 0$, the $\phi$ function becomes $\phi(t) = \ln(t)$ and its inverse is $\phi^{-1}(u) =\exp(u)$. In this case, the $\beta$-mean~(\ref{phi_beta_mean}) corresponds to the geometric mean:
\[\lim_{\beta\to 0} \overline{\Mb}_{\beta}  =\exp\left\{\frac{1}{m} \sum_{\ell=1}^{m}
\ln\big(\Mb^{(\ell)}\big) \right\}.\]
A detailed derivation of this limiting case is provided in Appendix~\ref{proof_limiting_g_mean}.
\end{rmk}

The central server aggregator proposed in Algorithm~\ref{algorithm_disPCA} is built upon the concept of the matrix $\beta$-mean (\ref{phi_beta_mean}) with $\Mb^{(\ell)}=\widehat{\Gammab}^{(\ell)}_{q} \widehat{\Lambdab}^{(\ell)}_{q} \widehat{\Gammab}^{(\ell)^{\top}}_{q}$.
 One advantage of using the matrix $\beta$-mean for aggregation is its robustness, stemming from its connection to the minimum matrix $\beta$-divergence estimation criterion. The density-based minimum $\beta$-divergence estimation for $\beta>0$ (Basu et al. 1998; Jones et al. 2001) has been shown to exhibit strong robustness properties in statistical inference. This connection will be further discussed in the next subsection.


\subsection{Connection to matrix $\betab$-divergence}\label{link_to_matrix_beta}

In many applications, such as non-negative matrix factorization and machine learning tasks, $\beta$-divergence offers a robust measure for handling various noise models and data distributions. It is flexible and retains general properties of the Bregman vector divergence (Bregman 1967), such as convexity and non-negativity.
In this section, we introduce a novel matrix $\beta$-divergence, and explore its connection to the matrix $\beta$-mean. Important properties of the Bregman matrix divergence are detailed in Appendix A of Kulis et al. (2009).
\begin{defn}[Matrix $\beta$-divergence]
Let $\Mb$, $\Mb_1$ and $\Mb_2$ be $p \times p$ matrices in $\cal P$, and $\Phi_{\beta} : \real^{p \times p}
\rightarrow \real $ be a strictly convex and differentiable function with its derivative $\phi_{\beta}(\cdot)$.
Consider the following Bregman generating function: for $\beta \in \real  \setminus\{0, -1\}$,
\begin{eqnarray}\label{generating_func}
 \Phi_{\beta}(\Mb)  =  \frac{1}{\beta (\beta+1)}  \tr \left( \Mb^{\beta+1} -(\beta+1)\Mb +\beta\Ib_{p}  \right)
\end{eqnarray}
with its derivative $\phi_{\beta}(\Mb) = \frac{1}{\beta}\left(\Mb^{\beta} - \Ib_{p} \right)$. 
The resulting Bregman matrix divergence is
\begin{eqnarray}\label{matrix_beta_div}
D_\beta(\Mb_{1},\Mb_{2})
&=&  \frac{1}{\beta(\beta+1)} \tr \left(\Mb_{1}^{\beta+1} +\beta\Mb_{2}^{\beta+1}
    -(\beta+1) \Mb_{2}^\beta \Mb_{1} \right), 
\end{eqnarray}
which we refer to as the matrix $\beta$-divergence. 
\end{defn}
\noindent To demonstrate that $\Phi$ in~(\ref{generating_func}) is a legitimate Bregman generating function, we will prove its strict convexity in Appendix~\ref{proof_convexity}, where we also provide the derivation for~(\ref {matrix_beta_div}). Note that 
the case of $\beta=1$ reduces to the Frobenius norm matrix distance.


In equation~(\ref{generating_func}) and for  a fixed $\beta \in \real  \setminus\{0, -1\}$, the second and third terms on the right-hand side of the equation will not affect the definition of matrix $\beta$-divergence, and thus, they can be removed. However, we retain these two terms here is to ensure the validity of $\Phi_\beta$ in the limiting cases for $\beta\to 0$ and $\beta\to -1$.
The limiting cases of $\beta$-divergence between two density functions were introduced by Cichocki and Amari (2010) and Eguchi and Komori (2022). Here, we extend the limiting cases to the matrix divergence. The extension is stated in the following proposition.

\begin{prop} [Limiting cases] \label{limiting_cases} The limiting cases of the matrix $\beta$-divergence are summarized below. \\
(i) We have $\lim_{\beta\to 0}
\Phi_{\beta}(\Mb)= \tr \left(\Mb \ln\left(\Mb \right) -\Mb \right)
+ p$, and
\begin{eqnarray}
\lim_{\beta\to 0} D_\beta(\Mb_{1},\Mb_{2}) &=&
\tr \left(\Mb_{1}\ln \left(\Mb_{1}\Mb_{2}^{-1}\right)-
\Mb_{1} + \Mb_{2} \right), \label{vonNeumann_div}
\end{eqnarray}
which is the von Neuman matrix divergence and is denoted as $D_{\rm vN}(\Mb_{1},\Mb_{2})$.\\
(ii) We have $\lim_{\beta\to -1} \Phi_{\beta}(\Mb) = -\ln \det (\Mb) +\tr\left(\Mb \right) - p$, and
\begin{eqnarray}
\lim_{\beta\to -1} D_\beta(\Mb_{1},\Mb_{2})
&=& \tr \left(\Mb_{1}\Mb_{2}^{-1}\right) - \ln \det(\Mb_{1}
    \Mb_{2}^{-1}) - p, \label{logdet_div}
\end{eqnarray}
which is the log-determinant matrix divergence and is denoted as $ D_{\rm ld}(\Mb_{1},\Mb_{2}) $.
\end{prop}
We now state the connection of matrix $\beta$-mean and matrix $\beta$-divergence. The following proposition demonstrates that the matrix $\beta$-mean exactly minimizes the matrix $\beta$-divergence in $\cal P$. 

\begin{prop}[Minimum matrix $\beta$-divergence estimation]\label{min_beta_div_est}
For $\Mb^{(\ell)}\in {\cal P}$, $\ell =1, 2, \ldots, m$,  the matrix $\beta$-mean in~(\ref{phi_beta_mean}) can be obtained by minimizing the matrix $\beta$-divergence:
\begin{eqnarray}
\overline{\Mb}_{\beta}   = \argmin_{\Mb\in \cal P }
\frac{1}{m} \sum_{\ell=1}^m D_{\beta}( \Mb,\Mb^{(\ell)}).
\end{eqnarray}
\end{prop}
\noindent Note that the matrix $\beta$-divergence is not symmetric in its two arguments. Typically, in a statistical divergence, the first argument represents the true, or the data distribution, and the second argument represents the model. However, in this article, we exchange their order. The rationale for this switch is explained in Remark~\ref{data_model} in the Appendix.

The aim of $\beta$-DPCA in aggregating local PCAs to form a final estimator for $\Gammab_{r}$ can be formulated as the minimum matrix $\beta$-divergence estimation given in Proposition~\ref{min_beta_div_est}. More specifically, given the target rank $r$ and the data matrices $\left\{\Mb^{(\ell)}\right\}_{\ell=1}^{m}$, the minimum matrix $\beta$-divergence estimation criterion aims to find a rank-$r$ model $\Mb_r:= \Gammab_{r}\Lambdab_r\Gammab^{\top}_{r}$ that best represents the center of these matrices in terms of minimizing divergence:
\begin{equation}\label{aggre_obj}
\argmin_{\Gammab_r, \Lambdab_r} \frac{1}{m} \sum_{\ell=1}^m {D_\beta(\Mb_r, \Mb^{(\ell)}  )},
\end{equation}
where $\Gammab_r$ is a $p\times r$ orthogonal matrix and $\Lambdab_r$ is a diagonal matrix consisting of~$r$ nonnegative diagonal elements. The following result connects the proposed $\beta$-DPCA to the minimum matrix $\beta$-divergence estimation criterion.

\begin{cor}\label{cor_min_beta_div_est}
The eigenvectors $\widehat{\Gammab}_{r}$ of $\beta$-DPCA minimizes the matrix $\beta$-divergence (\ref{aggre_obj}) subject to a regularization by $\delta \Ib_p$ when $\beta<0$.  
\end{cor}

\subsection{Stability of eigenvector ordering}\label{eig_perturb}

The aim of this section is to investigate the stability (or robustness) of $\beta$-DPCA in estimating $\Sr$ under eigenvalue perturbation. Specifically, we seek to understand how the magnitude of perturbation for a noise eigenvalue affects the ordering of eigenvectors. A larger perturbation in a noise eigenvalue indicates a higher chance that this perturbed eigenvalue will rank before some of the true signal eigenvalues, causing its corresponding eigenvector to become a leading eigenvector, and hence, a less accurate estimation of $\Sr$. To simplify the discussion, we assume that the eigenvectors are fixed and not perturbed. This assumption helps in focusing on the impact of eigenvalue perturbation alone, as considering perturbed eigenvectors would complicate the analysis significantly.

Since the eigenvector matrix $\Gammab$ is not perturbed, it suffices to consider the case of $\Mb^{(\ell)}=\Lambdab^{(\ell)}$ for all $\ell$. Consider a scenario where a local machine, say machine $m$, undergoes a perturbation in its eigenvalues by an amount $\Db$, where $\Db$ is a diagonal matrix with $d_l > 0$ for a certain $l > r$, while all other entries remain zero. The $\beta$-mean of the matrices is 
\begin{equation}\label{apply_prop}
\Lambdab_\beta:=\left\{ \frac1m \sum_{\ell=1}^{m-1} \left(\Lambdab^{(\ell)}\right)^\beta + \frac1m \left(\Lambdab^{(m)}+\Db\right)^\beta\right\}^{1/\beta} 
= \left\{ \big(\overline{\Lambdab}_\beta\big)^\beta + \frac1m \tilde\Lambdab^\beta\right\}^{1/\beta}, 
\end{equation}
where $\overline{\Lambdab}_\beta = \big\{\frac1m \sum_{\ell=1}^m \left(\Lambdab^{(\ell)}\right)^\beta\big\}^{1/\beta}$ and $\tilde\Lambdab$ is a diagonal matrix with $\tilde\lambda_l =\big\{ (\lambda_l ^{(m)}+d_l)^\beta -(\lambda_l ^{(m)})^\beta\big\}^{1/\beta}>0$, while all other entries are zero. With the eigenvalue perturbation, we aim to determine the maximum tolerance size of perturbation $\tau$, such that $\tilde{\lambda}_{l} < \tau$ if and only if 
the following condition holds:
\begin{equation}\label{eigenvalue_cond}
\mbox{(order invariance condition)}\quad \min_{j\leq r} \lambda_{\beta,j}  >  \max_{j'>r} \lambda_{\beta, j'},
\end{equation}
where $\lambda_{\beta,j}$ is the $j^\th$ diagonal element of $\Lambdab_\beta$. That is, the order invariance condition is satisfied if and only if the perturbation of 
$\tilde{\lambda}_{l}$ is controlled to be less than $\tau$. This ensures that the true leading $r$ eigenvectors remain ranked before the noise eigenvectors. In this situation, a consistent estimation of $\Sr$ is still achievable under the eigenvalue perturbation in~(\ref{apply_prop}).
A larger $\tau$ then indicates a greater tolerance to perturbation. We have the following results.

\begin{prop}\label{prop_ordering_perturb}
Assume the eigenvalue perturbation that leads to (\ref{apply_prop}). Let $\bar\lambda_j$ be the $j^\th$ diagonal element of $\overline{\Lambdab}_\beta$. We have the following results regarding $\beta$-DPCA.
\begin{itemize}
\item[(i)] 
For $\beta > 0$, $\tau = \left\{m(\bar\lambda_{r}^{\beta} - \bar\lambda_{l}^{\beta})   \right\}^{1/\beta}$.

\item[(ii)] 
For $\beta \to 0$, $\tau=  \left({\bar\lambda_{r}} / {\bar\lambda_{l}} \right)^m$.

\item[(iii)]
For $\beta < 0$, $\tau = \infty$ which indicates the order invariance property of $\beta$-DPCA under any perturbation.
\end{itemize} 
\end{prop}

The discussion of perturbation is presented under a simplified scenario, which, although not entirely realistic, provides valuable insights into the impact of perturbation.
This result demonstrates how a single local perturbation influences the aggregated eigenvalue structure. It also provides a measure of robustness in the ordering of eigenvectors under eigenvalue perturbation, reinforcing the stability of $\beta$-DPCA with $\beta < 0$.

\section{Simulation Studies}\label{numerical}

\subsection{Simulation settings}

For each simulation run, we generate $\Gammab$ by orthogonalizing a $p\times p$ random matrix consisting of iid $N(0,1)$ entries. As to $\Lambdab$, the signal eigenvalues are set to be $\lambda_{j} = 1 + (p/n)^{1/2} + p^{1/(1+j)}$ for $j \leq r$, and the noise eigenvalues $\{ \lambda_{j} \}_{j=r+1}^{p}$ are drawn from ${\rm Uniform}(0.5, 1.5)$. 
Given $\Sigmab=\Gammab\Lambdab\Gammab^\top$, the data matrix $\Xb$ is generated from either a mean-zero Gaussian distribution with covariance matrix $\Sigmab$ or a mean-zero multivariate $t$-distribution with 3 degrees of freedom and covariance matrix $\Sigmab$.
To implement $\beta$-DPCA, we consider $\mathcal{B}= \{-1, 0, 1\}$ ($\beta=0$ is understood as the limiting case $\beta\to 0$) as the candidate set of the tuning parameter $\beta$. The sampling dimension is set to $q=10$, and the regularization parameter is fixed at $\delta =  10^{-5}$.

The performance metric of an estimate $\hat\Gammab_k$ (or $\tilde\Gammab_k$) and the ground truth $\Gammab_r$ is evaluated by the average of canonical angles between  ${\rm span}(\hat\Gammab_k)$ (or ${\rm span}(\tilde\Gammab_k)$) and  ${\rm span}(\Gammab_r)$:
\begin{equation}
\rho_{k} = \frac{1}{r} \sum_{j=1}^{r} \sigma_{kj} \in [0, 1],~ k \geq r,
\end{equation}
where $\sigma_{kj}$ is the $j^\th$  singular value of
$\widehat{\Gammab}^{\top}_{k}\Gammab_{r}$ (or
$\tilde{\Gammab}^{\top}_{k}\Gammab_{r}$). Here, $k \ge r$ is
considered to assess the similarity of the ground truth with an
enlarged eigenspace. For $\rho_{k}=1$, it indicates that the
estimated eigenspace contains the true eigenspace, i.e., ${\rm
span}(\Gammab_r) \subseteq {\rm span}(\hat\Gammab_k)$ (or ${\rm
span}(\Gammab_r) \subseteq {\rm
span}(\tilde\Gammab_k)$), while $\rho_{k}=0$ indicates that the
two eigenspaces are perpendicular and have zero as the only common
element, i.e., ${\rm span}(\Gammab_r) \cap {\rm span}(\hat\Gammab_k) =\{0\}$ (or ${\rm span}(\Gammab_r) \cap {\rm span}(\tilde\Gammab_k) =\{0\}$). The average of $\rho_{k}$ for $k \in \{r, r+1, \ldots, 15\}$ based on 100 replicate runs are reported under $r=5$, $n=250$, $p\in\{500,1000\}$, and $m\in\{5,10\}$.

\subsection{Results}\label{sim_results}

\begin{figure}[h!]
\vspace{-1.5cm}

\subfigure[Multivariate Gaussian distribution.]{\parbox{\textwidth}{\centering
\mbox{\hspace{-0.6cm}
\includegraphics[scale=0.51]{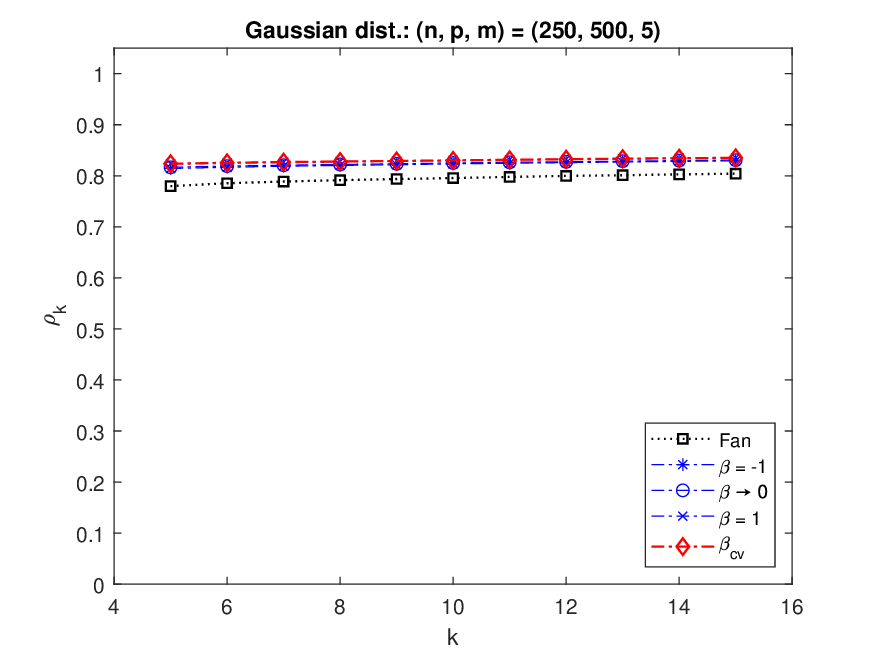}
\includegraphics[scale=0.51]{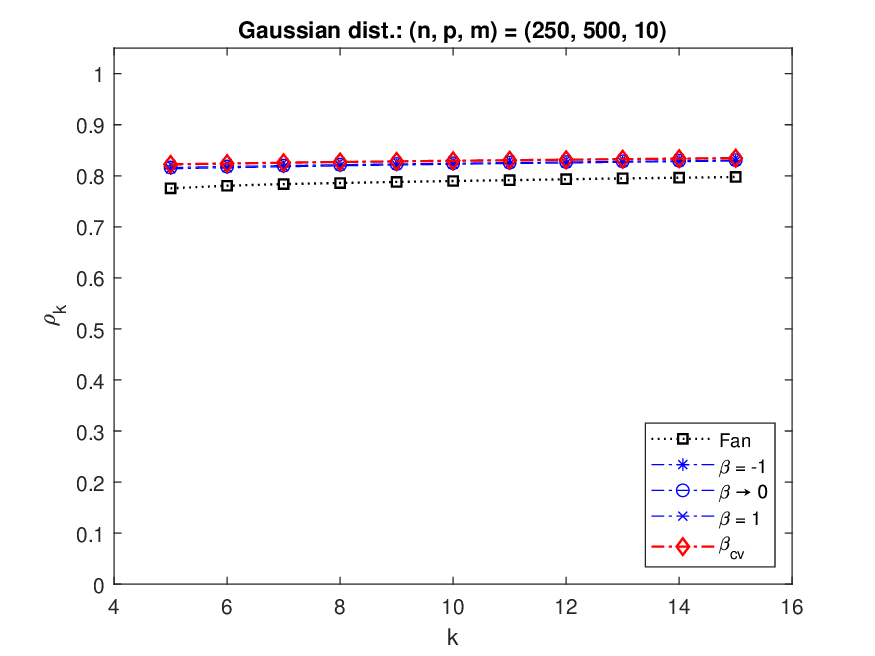}}

\mbox{\hspace{-0.6cm}
\includegraphics[scale=0.51]{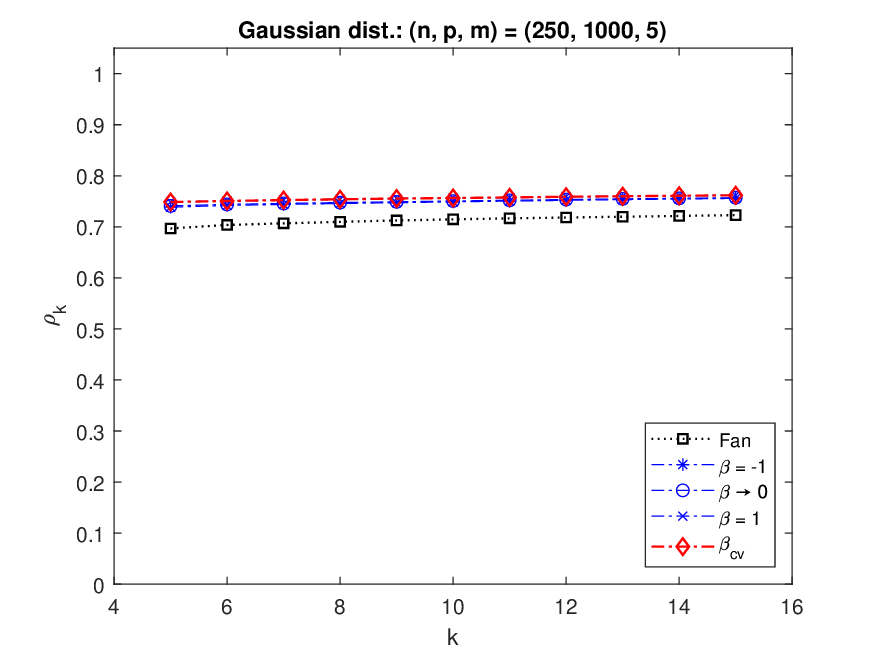}
\includegraphics[scale=0.51]{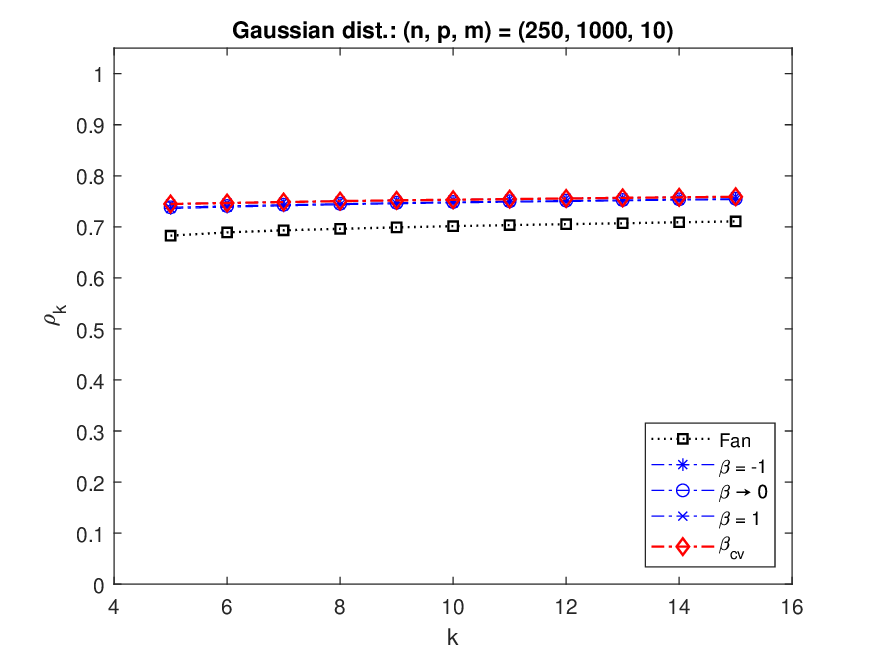}}}}

 \subfigure[Multivariate $t$-distribution with 3 degrees of freedom.]{\parbox{\textwidth}{\centering
\mbox{\hspace{-0.6cm}
\includegraphics[scale=0.51]{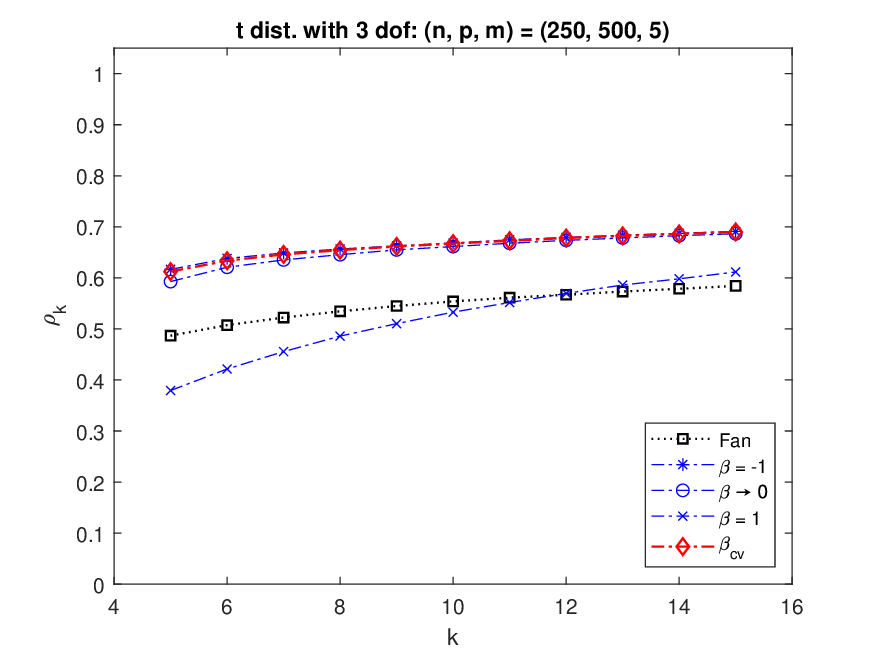}
\includegraphics[scale=0.51]{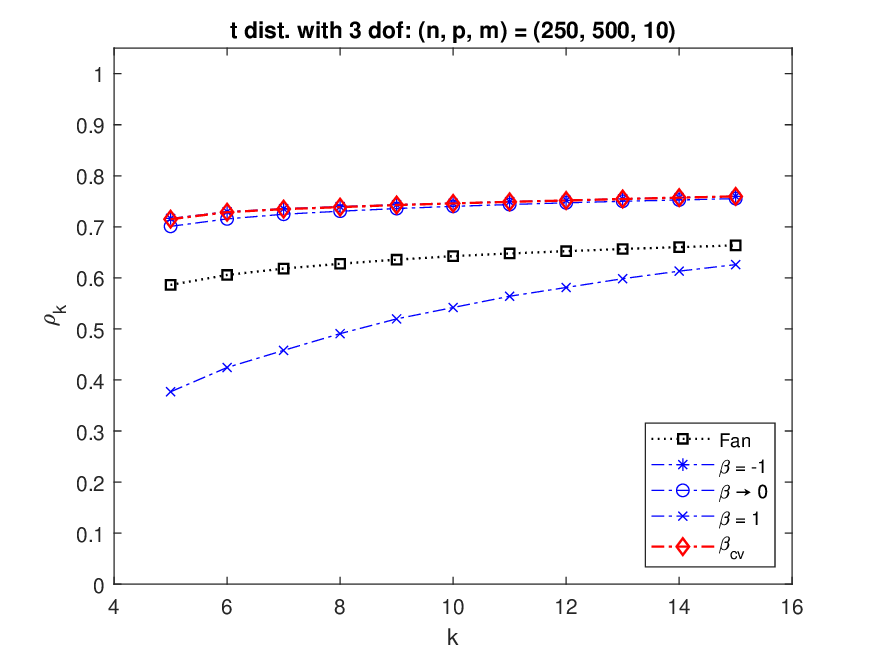}}

\mbox{\hspace{-0.6cm}
\includegraphics[scale=0.51]{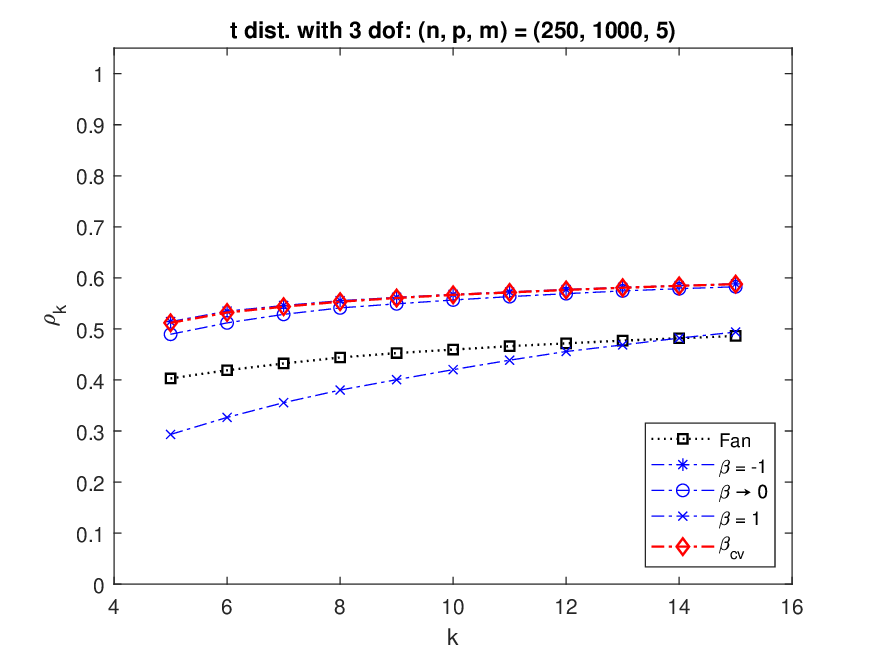}
\includegraphics[scale=0.51]{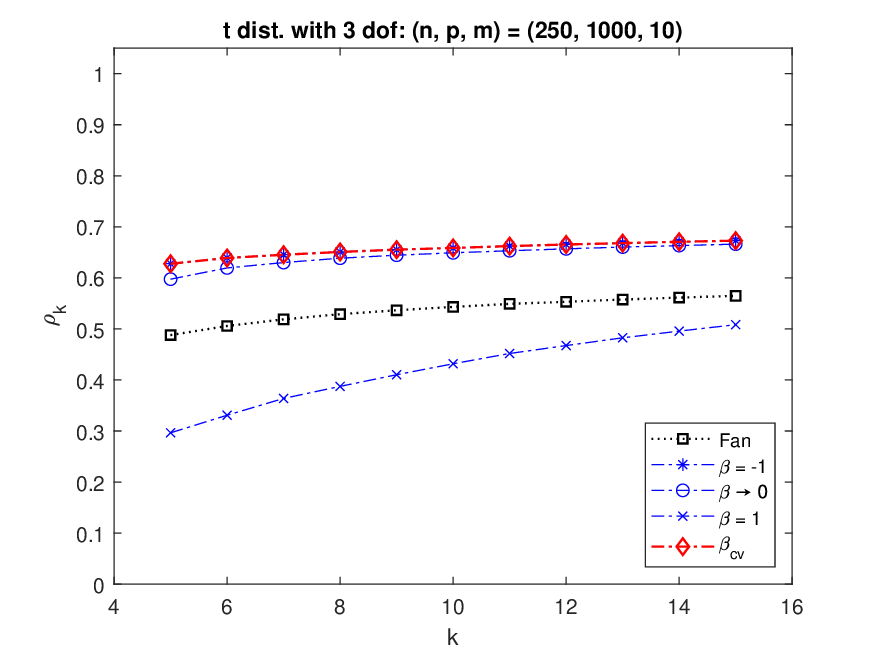}}}}

\caption{The mean similarity measure $\rho_{k}$ for $k\in\{ r, r+1, \ldots, 15\}$ under $r=5$, $n=250$, $p\in\{500,1000\}$ and $m\in\{5,10\}$.}\label{results_similarities}
\end{figure}

We compare the performance of our $\beta$-DPCA (including the cases of $\beta=\pm 1$,
$\beta\to 0$ and $\beta$ being selected by 5-fold CV) with the
DPCA algorithm of Fan et al. (2019). The frequencies of
the selected $\beta$ values by CV are reported in
Table~\ref{results_freq_beta}. The performance comparison for the multivariate Gaussian distribution and different combinations of $(p, m)$ is presented in Figure~\ref{results_similarities}(a).
For $p = 500$ and $1000$, $\beta$-DPCA demonstrates similar $\rho_{k}$ values across three distinct values of $\beta$. Moreover, our $\beta_{\rm{cv}}$ shows performance comparable to the best among the three $\beta$ values. Compared to Fan et al. (2019), our method performs slightly better. Also, there is no apparent difference when $m$ increases from $5$ to $10$.

The performance comparison for multivariate $t$-distribution with
3~degrees of freedom and  different combinations of $(p, m)$ is
presented in Figure~\ref{results_similarities}(b). 
The $\beta$-DPCA with $\beta=1$, which averages local PCA results as an approximation of classical PCA, performs the worst, as expected, since the classical PCA  is sensitive to heavy-tailed outliers.
However, for $\beta=-1$ and
$\beta\to 0$, $\beta$-DPCA outperforms the method by Fan et al.
(2019) due to the proper utilization of eigenvalue information and the dimension oversampling, which further enhance the accuracy of eigenspace estimation.
Our $\beta_{\rm{cv}}$ consistently selects the optimal
$\beta$, achieving performance comparable to the best performer.
Additionally, as $m$ increases, all methods show improvement.

\begin{table}[h!]
\caption{The frequencies of the selected $\beta$ value by 5-fold CV based on 100 replicate runs.}\label{results_freq_beta}
\begin{center}

\begin{tabular}{|c|ccc| ccc|}
\hline
& \multicolumn{3}{c|}{{\bf Gaussian}} &  \multicolumn{3}{c|}{ $\bm{t_3}$}\\

 $(p, m)$        & $\beta=-1$ &  $\beta\to 0$ &  $\beta= 1$    & $\beta=-1$ &  $\beta \to 0$ &  $\beta= 1$\\ \hline
 $( 500, 5) $    &  0  & 0   &  100   &  76  &  24  & 0 \\
 $( 500, 10) $  &  0  & 0   &  100   &  90  &  10  & 0 \\
 $( 1000, 5) $  &  0  & 0   &  100   &  88  &  12  &  0  \\
 $( 1000, 10) $ & 0   & 1    &  99     & 98   & 2  & 0 \\
 \hline
\end{tabular}
\end{center}
\end{table}

\clearpage

\section{Conclusion}\label{conclusion}

In this article, we developed a $\beta$-DPCA algorithm under a sample-partitioned setting. Our method introduces the concept of the matrix $\beta$-mean to provide a flexible and robust approach to incorporate the eigenvalue information into the aggregation step for estimating the PCA eigenspace. We also proposed a cross-validation procedure for selecting the optimal $\beta$ value, ensuring that the aggregation method is well-suited to different data characteristics. Additionally, we investigated the stability of eigenvector ordering under eigenvalue perturbation, demonstrating the robustness of $\beta$-DPCA, especially for the case of $\beta=-1$. Simulation studies reveal that $\beta$-DPCA with $\beta$ selected by cross-validation achieves more accurate eigenspace estimation compared to the conventional DPCA methods.

\section*{References}

\begin{description}

 \item 
Basu, A., Harris, I. R., Hjort, N. L., and Jones, M. C. (1998). Robust and efficient estimation by minimising a density power divergence. {\it Biometrika}, 85(3), 549-559.

 \item 
Bhaskara, A. and Wijewardena, P. M. (2019). On distributed
averaging for stochastic k-PCA. {\it Advances in Neural
Information Processing Systems}, 32, 11024-11033.

\item 
Bregman, L. M. (1967). The relaxation method of finding the common point of convex sets and its application to the solution of problems in convex programming. {\it USSR Computational Mathematics and Mathematical Physics}, 7(3), 200-217.

\item 
Cichocki, A. and Amari, S. I. (2010). Families of alpha-beta-and
gamma-divergences: Flexible and robust measures of similarities.
{\it Entropy}, 12(6), 1532-1568.

\item 
Dhillon, I. S. and Tropp, J. A. (2008). Matrix nearness problems with Bregman divergences. {\it SIAM Journal on Matrix Analysis and Applications}, 29(4), 1120-1146.

\item 
Eguchi, S. and Kato, S. (2010). Entropy and divergence associated
with power function and the statistical application. {\it
Entropy}, 12(2), 262-274.

\item 
Eguchi, S. and Komori, O. (2022). {\it Minimum Divergence Methods in
Statistical Machine Learning: From an Information Geometric
Viewpoint}. Springer.

\item 
Eguchi, S., Komori, O., and Ohara, A. (2018). Information geometry
associated with generalized means. In {\it Information Geometry
and Its Applications: On the Occasion of Shun-ichi Amari's 80th
Birthday}, IGAIA IV Liblice, Czech Republic, June 2016 (pp.
279-295). Springer International Publishing.

\item 
Fan, J., Wang, D., Wang, K., and Zhu, Z. (2019). Distributed estimation of principal eigenspaces. {\it Annals of Statistics}, 47(6), 3009-3031.

 \item 
Jones, M. C., Hjort, N. L., Harris, I. R., and Basu, A. (2001). A comparison of related density‐based minimum divergence estimators. {\it Biometrika}, 88(3), 865-873.

\item 
Kulis, B., Sustik, M. A., and Dhillon, I. S. (2009). Low-rank
kernel learning with Bregman matrix divergences. {\it Journal of
Machine Learning Research}, 10(2), 341-376.

\item 
Li, X., Wang, S., Chen, K., and Zhang, Z. (2021).
Communication-efficient distributed SVD via local power
iterations. In {\it International Conference on Machine Learning}
(pp. 6504-6514). PMLR.

\item 
Yata, K. and Aoshima, M. (2010). Effective PCA for high-dimension, low-sample-size data with singular value decomposition of cross data matrix. {\it Journal of Multivariate Analysis}, 101(9), 2060-2077.

\item  
Wu, S. X., Wai, H. T., Li, L., and Scaglione, A. (2018). A review of distributed algorithms for principal component analysis. {\it Proceedings of the IEEE}, 106(8), 1321-1340.

\end{description}

\end{document}